\documentclass[runningheads]{llncs}
\usepackage{graphicx}
\usepackage{comment}
\usepackage{amsmath,amssymb} % define this before the line numbering.
\usepackage{color}
\usepackage{url} 
\usepackage{hyperref}

\usepackage[labelfont=bf]{caption}

\newcommand\blfootnote[1]{%
  \begingroup
  \renewcommand\thefootnote{}\footnote{#1}%
  \addtocounter{footnote}{-1}%
  \endgroup
}

\usepackage{array}
\newcolumntype{L}[1]{>{\raggedright\let\newline\\\arraybackslash\hspace{0pt}}m{#1}}
\newcolumntype{C}[1]{>{\centering\let\newline\\\arraybackslash\hspace{0pt}}m{#1}}
\newcolumntype{R}[1]{>{\raggedleft\let\newline\\\arraybackslash\hspace{0pt}}m{#1}}

\begin{document}
% \renewcommand\thelinenumber{\color[rgb]{0.2,0.5,0.8}\normalfont\sffamily\scriptsize\arabic{linenumber}\color[rgb]{0,0,0}}
% \renewcommand\makeLineNumber {\hss\thelinenumber\ \hspace{6mm} \rlap{\hskip\textwidth\ \hspace{6.5mm}\thelinenumber}}
% \linenumbers
\pagestyle{headings}
\mainmatter
\def\ECCVSubNumber{1415}  % Insert your submission number here 

%\title{Deep Implicit Representations for Humans with Applications to Single View 3D  Reconstruction and Novel View Synthesis} % Replace with your title 
%\title{Deep Implicit Representations for Free-Viewpoint Rendering of Humans From a Singe Image}
\title{Neural Re-Rendering of Humans from a \\Single Image} 

% CAMERA READY SUBMISSION
%\begin{comment}
% \titlerunning{Neural Human Re-Rendering} 
\titlerunning{Neural Re-Rendering of Humans from a \\Single Image} 
% If the paper title is too long for the running head, you can set
% an abbreviated paper title here
%%Second Author\inst{2,3}\orcidID{1111-2222-3333-4444} \and
%\author{Kripasindhu Sarkar\inst{1} \and
%Dushyant Mehta\inst{1} \and
%Weipeng Xu\inst{2} \and \\
%Vladislav Golyanik\inst{1} \and
%Christian Theobalt\inst{1}}
\author{Kripasindhu Sarkar$^1$ \hspace{15pt}  Dushyant Mehta$^1$ \hspace{15pt} Weipeng Xu$^2$ \vspace{3pt}\\ 
Vladislav Golyanik$^1$ \hspace{15pt} Christian Theobalt$^1$
}

\authorrunning{Sarkar \textit{et al.}}
% First names are abbreviated in the running head.
% If there are more than two authors, 'et al.' is used.
%
% \institute{Max Planck Institute for Informatics, Saarland Informatics Campus \and Facebook Reality Labs}   

\institute{$^1$MPI for Informatics, SIC \hspace{20pt} $^2$Facebook Reality Labs}

%\end{comment}
%******************
\maketitle

\begin{abstract}
Human re-rendering from a single image is a starkly underconstrained problem, and state-of-the-art algorithms often exhibit undesired artefacts, such as over-smoothing, unrealistic distortions of the body parts and garments, or implausible changes of the texture. 
To address these challenges, we propose a new method for neural re-rendering of a human under a novel user-defined pose and viewpoint,  given one input image. 
Our algorithm represents body pose and shape as a parametric mesh which can be reconstructed from a single image and easily reposed.
Instead of a colour-based UV texture map, our approach further employs a learned high-dimensional UV feature map to encode appearance.
This rich implicit representation captures detailed appearance variation across poses, viewpoints, person identities and clothing styles better than learned colour texture maps. 
The body model with the rendered feature maps is fed through a neural image translation network that creates the final rendered colour image.
The above components are combined in an end-to-end-trained neural network architecture that takes as input a source person image and images of the parametric body model in the source pose and desired target pose. 
Experimental evaluation demonstrates that our approach produces higher-quality single image re-rendering results than existing methods. 
%\dots
\end{abstract}

\section{Introduction} 

%\blfootnote{Project webpage:  \url{http://gvv.mpi-inf.mpg.de/projects/NeuralHumanReRendering/}}
\blfootnote{Project webpage:  \url{gvv.mpi-inf.mpg.de/projects/NHRR/}}
Algorithms to realistically render dressed humans under controllable poses and viewpoints are essential for character animation, 3D video, or virtual and augmented reality, to name a few. 
Over the past decades, computer graphics and vision have developed impressive methods for high-fidelity artist-driven and reconstruction-based human modelling, high-quality animation, and photo-realistic rendering.
However, these often require sophisticated multi-camera setups, and deep expertise in animation and rendering, and are thus costly, time-consuming and difficult to use.  
\begin{figure}[t!]
\begin{center}
    \includegraphics[width=\linewidth]{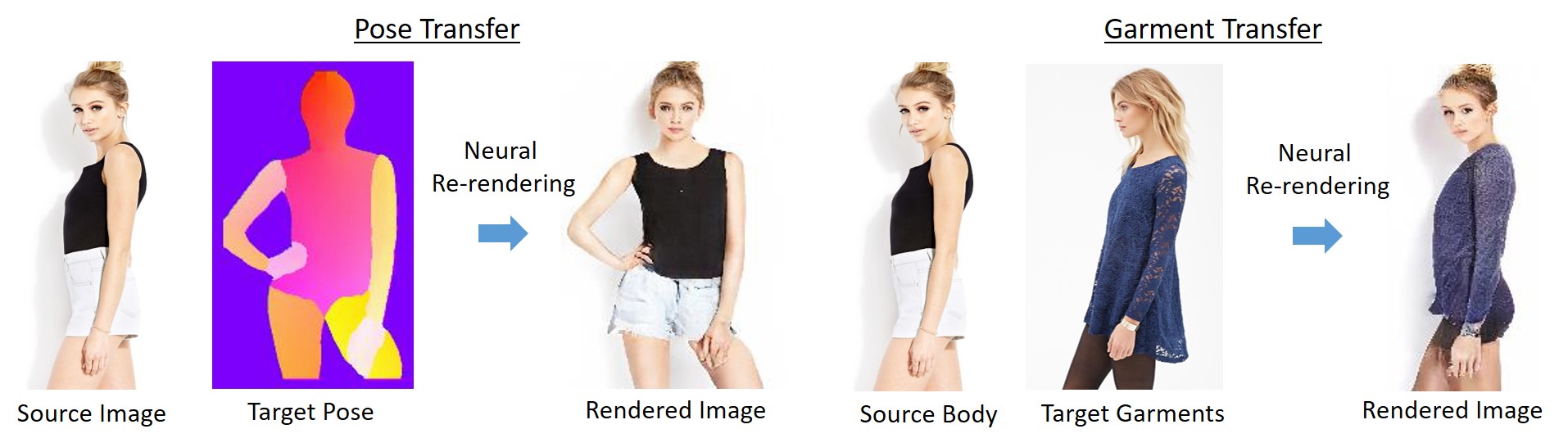}
\end{center}
\vspace{-0.7cm}
    \caption{Given an image of a person, our neural re-rendering approach allows synthesis of images of the person in different poses, or with different clothing obtained from another reference image.} 
    \label{fig:teaser} 
\vspace{-0.1cm}
\end{figure}
Recent advances in monocular human reconstruction and neural network-based image synthesis open up a radically different approach to the problem, neural re-rendering of humans from a single image.   
Given a single reference image of a person, the goal is to synthesise a photo-real image of this person in, for instance, a user-controlled new pose, modified body proportions, the same or different garments, or a combination of these. 

There has been tremendous progress in monocular human capture and re-rendering~\cite{Neverova2018,Liu2019,Siarohin_2019_NeurIPS,alldieck2019tex2shape,Lazova2019360DegreeTO,Grigorev2019CoordinateBasedTI,Kim2018,chan2019dance,lwb2019} towards this goal. 
However, owing to the starkly underconstrained nature of the problem, true photo-realism under all possible conditions has not yet been achieved.
Methods frequently exhibit unwanted over-smoothing and a lack of details in the rendered image, unrealistic distortions of body parts and garments, or implausible texture alterations.

We, therefore, propose a new algorithm for monocular neural re-rendering of a dressed human under a novel user-defined pose and viewpoint, which has starkly improved visual quality, see Figs.~\ref{fig:teaser}, \ref{fig:soa}, \ref{fig:garment}, \ref{fig:fashion}. 
We take inspiration from recent work on neural rendering of general scenes 
with a continuous \cite{Sitzmann2019} or a multi-dimensional feature representation with implicit~\cite{Thies2019} or explicit~\cite{sitzmann2019deepvoxels} occlusion handling that are learned from multi-view images or videos. 
Our algorithm represents body pose and shape with the SMPL parametric human surface model \cite{SMPL:2015},
%which can be reconstructed from a single image using the DensePose algorithm~\cite{Guler2018DensePose}, 
which can be easily reposed. 
% \CT{and reshaped ? - maybe add to supp mat.}. 
%based on semantically meaningful pose parameters.
%
Instead of modelling appearance as explicit colour maps, \textit{e.g.,} learned colour-based UV texture maps on the body surface \cite{Neverova2018,Grigorev2019CoordinateBasedTI}, we employ a learned high-dimensional UV feature map to encode appearance.
This rich implicit representation learns the detailed appearance variation across poses, viewpoints, person identities and clothing styles. 
%better than learned color texture maps. 
%
% 
Given a single image of a person, we predict pixel correspondences to the SMPL \cite{SMPL:2015} mesh using DensePose \cite{Guler2018DensePose}. 
We then extract partial UV texture maps based on the observed body regions and use a neural network to convert it to a complete UV feature map, with a $d$-dimensional feature per texel. 
The UV feature map is then rendered in the desired target pose and passed through a neural image translation network that creates the final rendered image. 
These components are combined in an end-to-end trained neural architecture. 
In quantitative experiments and a user study to judge the qualitative results, we show that the visual quality of our results improves over the current state of the art. 

\noindent\textbf{Contributions}. 
To summarise, our contributions are as follows: 
\begin{itemize} 
 \item A new end-to-end trainable method that combines monocular parametric 3D body modelling, 
 %a learned neural detail-preserving surface feature representation
a learned detail-preserving neural-feature based body appearance representation, and a neural network based image-synthesis network to enable highly realistic human re-rendering from a single image; 
 % \item a
 \item state-of-the-art results on the DeepFashion dataset \cite{Liu2016DeepFashion} which are confirmed with quantitative metrics, and qualitatively with a user study. 
\end{itemize}

%!TEX root = ../article.tex
\section{Related Work}

While our proposed approach relates to many sub-fields of visual computing, for brevity we only elaborate on the immediately relevant work on human body re-enactment and neural rendering methods for object and scene rendering. 

\subsection{Classical Methods for Novel View Synthesis} 
% of Scenes and Objects 
% 
Earlier methods for image-based 3D reconstruction and novel view synthesis rely on traditional concepts of multi-view geometry, explicit 3D shape and appearance reconstruction, and classical computer graphics or image-based rendering. 
Methods based on light fields use ray space representations or coarse multi-view geometry models for novel view synthesis \cite{Levoy1996,Gortler1996,Buehler2001}. 
To achieve high quality, dense camera arrays are required, which is impractical. 
Other algorithms capture and operate on dense depth maps \cite{Zhang2003}, layered depth images \cite{Shade1998}, 3D point clouds \cite{Agarwal2009C_ACM,Liu2010TVCG,schonberger2016structure}, meshes \cite{Matsuyama2004,Tung2009}, or surfels \cite{Pfister2000,Carceroni2002,Waschbuesch2005} for dynamic scenes. 
Multi-view stereo can be combined with fusion algorithms operating with implicit geometry and achieving more temporally consistent reconstructions over short time windows \cite{Dou2016,OrtsEscolano2016,Guo2017}. 
Dynamic scene capture and novel view synthesis were also shown with a low number of RGB or RGB-D cameras \cite{Yu2017,DoubleFusion2018,Huang2018,Yu2019}. 
While reconstruction is fast and feasible with fewer cameras, the coarse approximate geometry often leads to rendering artefacts. 

\subsection{Neural Rendering of Scenes and Objects} 
Recently, neural rendering approaches have shown promising results for scenes and objects. 
Image-based rendering (IBR) methods reconstruct scene geometry with classical techniques and use it to render novel views \cite{Debevec1998,Chaurasia2013}. 
Lack of observations can cause high uncertainty in novel views. 
On the other hand, neural rendering approaches~\cite{Sitzmann2019,sitzmann2019deepvoxels,thies2020imageguided,Zhu2018} can generate higher-quality results by leveraging collections of training data. 
Many applications of neural rendering have been recently shown, ranging from synthesising view-dependent effects \cite{Zhu2018,thies2020imageguided} to learning the shape and appearance priors from sparse data \cite{saito2019pifu,xu2019deepviewsynthesis}. 

Only a few works on neural scene representation and rendering can handle dynamic scenes ~\cite{Kim2018,lombardi2019neural}. 
% \unsure{which ones ??}
% 
% 
Some methods combine explicit dynamic scene reconstruction and traditional graphics rendering with neural re-rendering \cite{MartinBrualla2018,Kim2018,Kim2019Neural,thies2020imageguided}. 
Thies \textit{et al.}~\cite{Thies2019} combine \textit{neural textures} with the classical graphics pipeline for novel view synthesis of static objects and monocular video re-rendering. 
Their technique requires a scene-specific geometric proxy which has to be reconstructed before the training. 
Instead of more complex joint reasoning of the geometry and appearance needed from the intermediate representation by neural rendering approaches such as that of Sitzmann \textit{et al.}~\cite{Sitzmann2019}, 
for our human-specific application scenario the coarse geometry is handled by the posable SMPL mesh, with a feature map similar to the Thies \textit{et al.}~\cite{Thies2019} capturing clothing appearance, which includes fine-scaled geometry, and clothing textures. 

Several approaches address related problems such as generating images of humans in new poses \cite{Zhao2017,Balakrishnan2018,Ma18,Neverova2018,Pandey2019}, or body re-enactment from monocular videos \cite{chan2019dance}, which are discussed next. 

\subsection{Human Re-enactment and Novel View Rendering}

Recent work on photo-realistic human body re-enactment and novel view rendering can be sub-classified along various dimensions.

Object-agnostic approaches \cite{Siarohin_2019_NeurIPS,Siarohin_2019_CVPR} model deformable objects directly in the image space. 
Siarohin \textit{et al.}~\cite{Siarohin_2019_NeurIPS} learn keypoints in a self-supervised manner and capture deformations in the vicinity of the keypoints using affine transforms. 
Features extracted from the source image are deformed to the target using the predicted transformations and passed on to a generator. % network. 
Additional predictions of dis-occluded regions indicate to the generator the regions which have to be rendered based on the context. 
% which should 
%
Zhu {et al}.~\cite{Zhu2018extrapolation} leverage geometric constraints and optical flow for synthesising novel views of humans from a single image. 

Object-specific techniques have the same core components as above, \textit{i.e.,} colour or feature transformation from source to target, occlusion reasoning or inpainting, and photo-realistic image generation from the warped feature or colour image. 
The key difference is that the feature transformation, occlusion reasoning, and inpainting are guided by an underlying object model, which, in our case, is a parametric human body mesh.
Kim \textit{et al.}~\cite{Kim2018} achieve full control over the head pose and facial expressions in photo-realistic renderings of a target actor by an adversarial training with a performance of the target actor. 
DensePose Transfer \cite{Neverova2018} uses direct texture transfer from the input image to the SMPL model, inpaints the occluded regions of the texture and renders it in a new pose. 
This image is blended with the image resulting from direct conditional generation from the input image, input Densepose, and target Densepose. 
Zablotskaia \textit{et al.}~\cite{Zablotskaia2019DwNetDW} generate subsequent video frames of human motion and use a direct warping guided by the reference frame, previously generated frame, and DensePose representations of the past and future frames. 
Their method does not rely on an explicit UV texture map. 
ClothFlow \cite{Han_2019_ICCV} implicitly captures the 
geometric transformation between the source and target image by estimating dense flow. 
Chan\textit{et al.}~\cite{chan2019dance} learn a subject-specific puppeteering system using a video of the subject such that all parts of the subject's body are seen in advance. 
The GAN-based rendering is driven by 2D pose extracted from the target subject. 
Zhou \textit{et al.}~\cite{Zhou2019DanceDG} also learn a personalised model using piecewise affine transforms of the part-segmented source image for modelling pose changes, generating the person image in front of a clean {background} plate, with a second stage fusing a given {background} image with the generated person's image. 
%
% Different from the approach of 
In contrast to Liu \textit{et al.}~\cite{lwb2019}, we transfer appearance % information 
from source to target image using a UV feature map. 
Instead of directly predicting missing regions of the UV texture map, 
coordinate-based inpainting \cite{Grigorev2019CoordinateBasedTI} predicts correspondence between all regions on the UV texture map and the input image pixels. 
This results in more texture details in body regions that become dis-occluded when re-posing. 
As shown in Sec.~\ref{sec:results}, our UV feature map based approach yields results of much higher quality in comparisons. 
%
% Our method contrasts to 
Shysheya \textit{et al.}~\cite{Shysheya2019} explicitly model the body texture and implicitly handle the shape. 
In contrast, while we explicitly handle the coarse shape, we use a UV feature map to model the fine-scaled shape and clothing texture implicitly. % handle 
Lazowa \textit{et al.}~\cite{Lazova2019360DegreeTO} propose an approach for reconstruction of textured 3D human models from a single image. 
Similar to our approach, it extracts a partial UV texture map using DensePose but inpaints the UV texture map using a GAN based supervision directly applied to the texture map. 
Additionally --- and similar to Alldieck \textit{et al.}~\cite{alldieck2019tex2shape} --- it predicts a displacement map on top of the SMPL mesh to capture clothing details not present in the SMPL model. 
Our approach does not explicitly model clothing details. 

In contrast to existing methods, we propose a new end-to-end trainable method that combines monocular parametric 3D body  modeling~\cite{Neverova2018,Grigorev2019CoordinateBasedTI}, a learned neural detail-preserving surface feature representation~\cite{Thies2019}, and a neural image-synthesis network for highly realistic human re-rendering from a single image. % to enable 

%!TEX root = ../article.tex
\section{Method}
\label{sec:method}

Given an image $I_{s}$ of a person, we synthesise a new image of the person in a different target body pose. 
%We use SMPL UV parameterization \cite{TODO} to represent the human body texture. Instead of handcrafting or learning color texture from the image, we learn $d$ dimensional neural texture.
Our approach comprises of four distinct steps. The first step uses DensePose~\cite{Guler2018DensePose} to predict dense correspondences between the input image $I_{s}$ and the SMPL model. This allows a \textit{UV texture map} $T_{s}$ to be extracted for the visible regions. 
The second step uses a U-Net~\cite{RFB15a} based network, which we term \emph{FeatureNet}, to construct the full \textit{UV feature map} $F_{s}$ from the partial RGB UV texture map $T_{s}$. $F_{s}$ contains a $d$-dimensional feature representation for all texels, both visible and occluded in the source image. 
The third step takes a target pose as input, and `renders' the UV feature map $F_{s}$ to produce a $d$-dimensional \textit{feature image} $R_{s\rightarrow t}$. 

\begin{figure}[t!]
    \includegraphics[width=\linewidth]{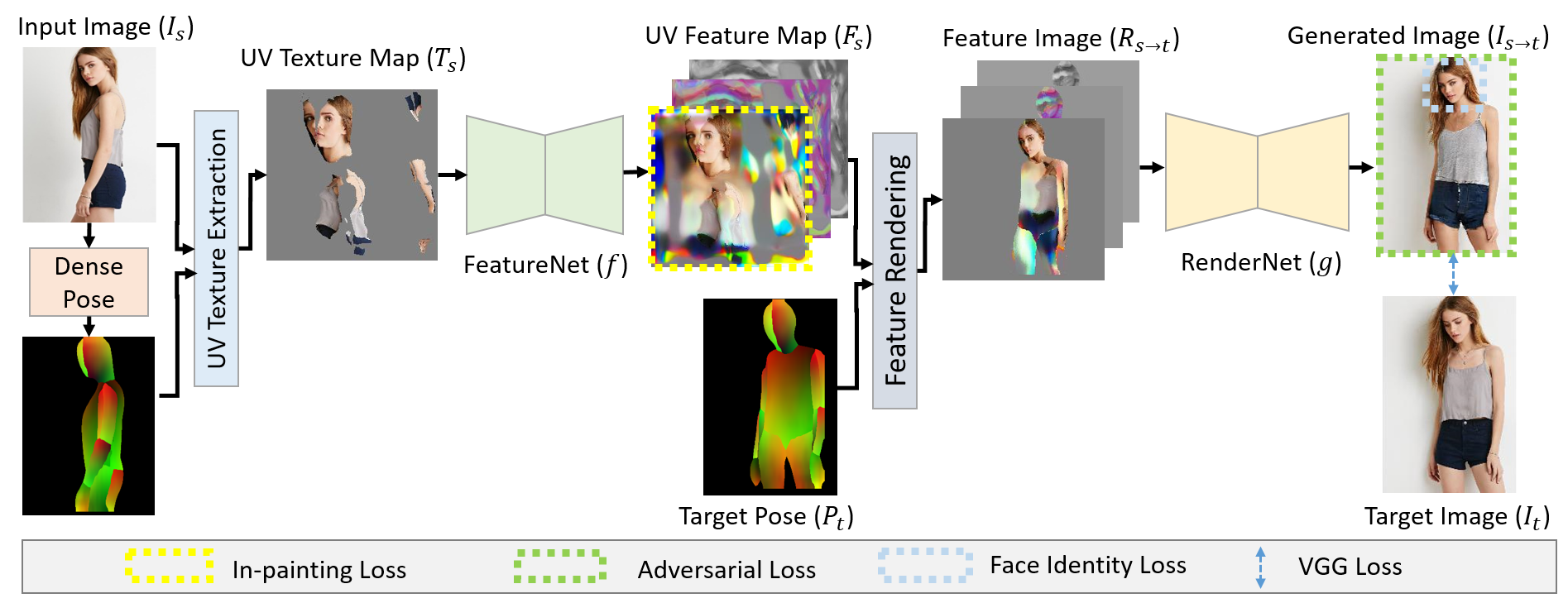}
    \caption{\textbf{Pipeline Overview}: Given a source image $I_s$, we extract the UV texture map $T_s$ of an underlying parametric body mesh model for the body regions visible in the image. 
    FeatureNet converts the partial UV texture map to a full UV feature map, which encodes a richer 16D representation at each texel. 
    Given a new pose $P_t$, the parametric body mesh can be re-posed and textured with the UV feature map to produce an intermediate feature image $R_{s\rightarrow t}$. 
    RenderNet converts the intermediate 16-channel feature image to a realistic image.}
    \label{fig:pipeline} 
\end{figure}

The fourth step uses a generator network based on Pix2PixHD~\cite{wang2018pix2pixHD}, which we term \emph{RenderNet}, to generate a photorealistic image $I_{s\rightarrow t}$ of the reposed person, from the input Feature image. The overview of our pipeline is shown in Fig.~\ref{fig:pipeline}.

\subsection{Extracting a Partial UV Texture Map from the Input Image} 

The pixels of the input image are transformed into UV space through matches predicted with DensePose. 
We use the ResNet-101 based variant of DensePose for predicting the correspondences for the body regions visible in the image. 
%computing DensePose of the input. 
The network is pre-trained on COCO-DensePose dataset and provides $24$ body segments and their part-specific \textit{U,V} coordinates of SMPL model. 
For easier mapping, the 24 part-specific UV maps are combined to form a single UV Texture map $T_s$ in the format provided in SURREAL dataset \cite{varol17_surreal} through a pre-computed lookup table. 
Note that one could putatively use monocular 3D pose estimation methods  (\textit{e.g.,}~\cite{hmrKanazawa17}) to compute SMPL parameters, and subsequently, the DensePose of the input image. 
However, frequent misalignments of the predictions with the end-effector positions in the image lead to significant artefacts in the UV texture map for hands and feet in that case and thus such an approach is not advised~\cite{alldieck2019tex2shape}.

\subsection{Generating the Full UV Feature Map} 
The partial (on account of occlusion) texture map $T_{s}$ is converted to a full UV feature map $F_{s}$ using a U-Net-like convolutional network $f$, which we term \emph{FeatureNet}. That is, 
\begin{equation}
F_s = f(T_s). 
\end{equation}
RenderNet comprises of four down-sampling blocks followed by four up-sampling blocks. 
Therefore, a partial input texture of the spatial dimension of $256 \times 256$ is transformed into a spatial dimension of $16 \times 16$ in the middle-most layer. 
Each downsampling block consists of two convolutions followed by maxpool operation. 
For up-sampling blocks, we use bilinear upsampling followed by two  convolutions. 
The final convolutional layer produces a $d$-dimensional (channel) \textit{UV feature map} which is used subsequently for rendering a feature image. 
The first three channels of the UV feature map can be supervised to in-paint the input partial UV texture map $T_{s}$,
%to represent an average colour of the object, \textit{i.e.,} 
thus having a small subset of the feature channels resemble the classical colour texture map. 
Our experiments use 16 feature channels.
% 

%\subsection{Re-posing Feature-map}
\subsection{Intermediate Feature Image Rendering} 
The SMPL mesh can be reposed using a target pose $P_{t}$, which can be extracted from a target image $I_{t}$, or obtained from a different source.
%, can be used to repose the SMPL mesh. 
% 
In our case, when given a target image $I_{t}$, we directly obtain the DensePose output, which is equivalent to the reposed SMPL model. 
Given the source feature map $F_s$, we render the SMPL mesh through the DensePose output $P_{t}$ to produce a $d$-dimensional \textit{feature image} $R_{s\rightarrow t}$. That is, 
\begin{equation}
R_{s\rightarrow t} = r(F_s, P_{t}).
\end{equation}
Note that this feature rendering operation $r$ can be conveniently implemented by differentiable sampling. 
In our experiments, we use bilinear sampling for this operation. 
The feature image $R_{s\rightarrow t}$, which captures the target pose and the source appearance is then used as input to the subsequent translation network.

\subsection{Creating a Photo-Realistic Rendering} 

In the final step, the feature image $R_{s\rightarrow t}$ is translated to a realistic image $I_{s\rightarrow t}$ using a translation network $g$ similar to Pix2Pix, which we term \emph{RenderNet}: % That is, 

\begin{equation}
I_{s\rightarrow t} = g(R_{s\rightarrow t}).
\end{equation}

%$$I_{s\rightarrow t} = g(R_{s\rightarrow t}).$$
% 
RenderNet comprises of (a) 3 down-sampling blocks, (b) 6 residual blocks, (c) 3 up-sampling blocks and finally (d) a convolution layer with Tanh activation that gives the final output. The discriminator for adversarial training of RenderNet also uses the multiscale design of Pix2PixHD \cite{wang2018pix2pixHD}. In our experiments, we use a three scale discriminator network for adversarial training. 

\subsection{Training Details and Loss Functions} 
\label{sec:lossfunctions} 
During training, our system takes pairs of images ($I_s$, $I_t$) of the same person (but in different poses) as input. 
Partial texture $T_s$ extracted from the source image $I_s$ is passed through the above-mentioned operations to produce the generated output $I_{s\rightarrow t}$. 
That is, 
\begin{equation}
%I_{s\rightarrow t} = g \circ r \circ f (T_s, P_{t}).
I_{s\rightarrow t} = g(r(f(T_s), P_{t})).
\end{equation}

%$$I_{s\rightarrow t} = g \circ r \circ f (T_s, P_{t}).$$

Note that all operations $g, r$ and $f$ are differentiable. 
We train the entire system \textit{end-to-end} and optimise the parameters of FeatureNet ($g(\cdot)$) and RenderNet ($f(\cdot)$). 
We use the combination of the following loss functions in our system: 
\begin{itemize}
\item \textbf{Perceptual Loss}: We use a perceptual loss based on the VGG network \cite{johnson2016perceptual} --- the difference between the activations on different layers of the pre-trained VGG network \cite{simonyan2014very} applied on the generated image $I_{s\rightarrow t}$ and ground-truth image target image $I_t$.
\begin{equation} 
L_p = \sum \frac{1}{N^j}|p^j(I_{s\rightarrow t}) - p^j(I_t)|.
\end{equation} 
Here, $p^j$ is the activation and $N^j$ the number of elements of the j-th layer in the VGG network pre-trained on ImageNet. 
\item \textbf{Adversarial Loss}: We use a multiscale discriminator $D$ of Pix2PixHD \cite{wang2018pix2pixHD} for enforcing adversarial loss $L_{adv}$ in our system. 
The multiscale discriminator $D$ is conditioned on both the generated and rendered feature images. 

\item \textbf{Face Identity Loss}: We use a pre-trained network to ensure that the extracted UV feature map and RenderNet preserve the \textit{face identity} on the cropped face of the generated and the ground-truth image: 
\begin{equation} 
L_{face} = |N_{face}(I_{s\rightarrow t}) - N_{face}(I_t)|. 
\end{equation} 
Here, $N_{face}$ is the pre-trained SphereFaceNet \cite{liu2017sphereface}

\item \textbf{Intermediate In-painting Loss}: To mimic classical colour texture map, we enforce a loss $L_{tex}$ on the first three channels of the output of the in-painting network. 
This loss is set to the sum of 1) $l_1$ distance of the visible part of the partial source texture and generated texture and 2) $l_1$ distance of the visible part of the partial target texture and generated texture. 
\end{itemize}
The final loss on the generator is then 
\begin{equation}L_G = \lambda_{VGG}L_p + \lambda_{face}L_{face} + \lambda_{tex}L_{tex} + \lambda_{GAN}L_{adv}. 
\end{equation} 
The conditional discriminator $D$ is updated every step enforcing binary cross-entropy loss on real and fake images. 
We train the networks end-to-end using Adam optimiser \cite{adam} with an initial learning rate of 2$\times 10^{-4}$, $\beta_1$ as 0.5 and no weight decay. 
The loss weights are set empirically to $\lambda_{GAN} = 1, \lambda_{VGG} = 10, \lambda_{face} = 5, \lambda_{tex} = 1$. For speed, we pre-compute DensePose on the images and directly read them as input. 

During testing, the system takes as input a single image of a person and a target DensePose. 
 The target pose can be extracted by DensePose RCNN on the image of the source person in a different pose (used in the experiments on DeepFashion dataset), or alternatively it can be obtained by reposing the SMPL mesh of the source body. In many cases, the actor can be a completely different person (see Figs.~\ref{fig:extreme_pose} and \ref{fig:fashion}). 
The neural texture is then rendered using the given target Densepose which is followed by the translation network to generate a realistic image of the source person in the target pose. 
% 

%!TEX root = ../article.tex
\section{Experimental Results} 
\label{sec:results}

\subsection{Experimental Setup}
\subsubsection{Datasets.}
We use the \textit{In-shop Clothes Retrieval Benchmark} of DeepFashion dataset \cite{Liu2016DeepFashion} for our main experiments. The dataset comprises of 52,712 images of fashion models with 13,029 different clothing items in different poses. For training and testing, we consider the split provided by Siarohin \textit{et al.}~\cite{Siarohin2019AppearanceAP}, which is also used by other related works \cite{Neverova2018,Grigorev2019CoordinateBasedTI}. 
We also show qualitative results of our method with Fashion dataset \cite{Zablotskaia2019DwNetDW}. Fashion dataset has 500 training and 100 test videos, each containing roughly 350 frames. 
%Each video contains a single person walking in a different pose.
The videos are single person sequences, containing different people catwalking in different clothes.

\begin{figure}[t]
    \includegraphics[width=\linewidth]{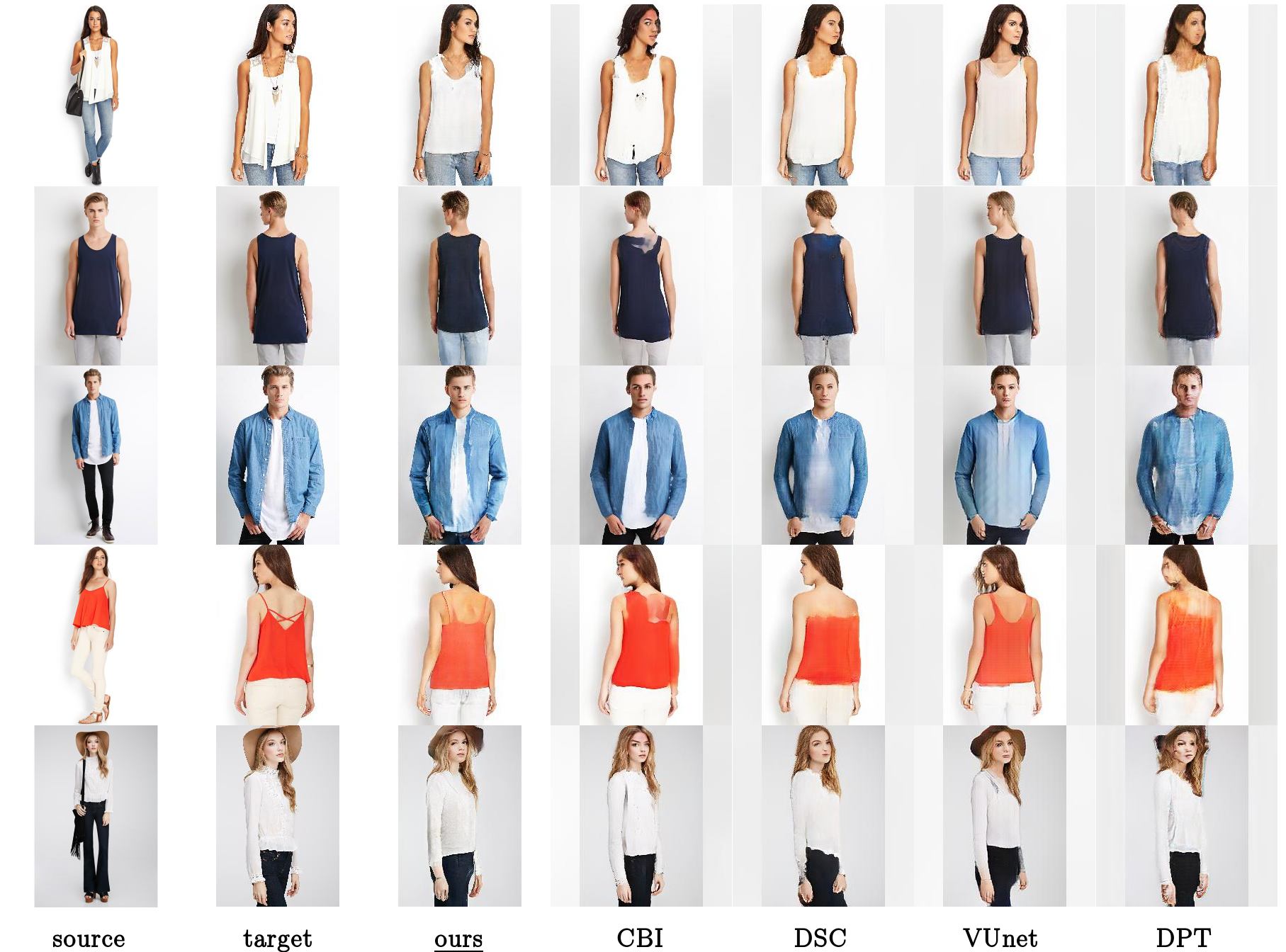}
    \caption{Results of our method,   % Coordinate Based Inpainting 
    CBI \cite{Grigorev2019CoordinateBasedTI},
    % Deformable GAN 
    DSC \cite{Siarohin2019AppearanceAP}, 
    % Variational Unet 
    VUnet \cite{esser2018variational} and 
    % Dense Pose Transfer 
    DPT \cite{Neverova2018}. 
    Our approach produces higher-quality renderings than the competing methods.} 
    \label{fig:soa} 
\end{figure}

\subsection{Comparison with the State of the Art}
We compare our results with four state-of-the-art methods, \textit{i.e.,}  Coordinate Based Inpainting (CBI) \cite{Grigorev2019CoordinateBasedTI}, Deformable GAN (DSC) \cite{Siarohin2019AppearanceAP}, Variational U-Net (VUnet) \cite{esser2018variational} and Dense Pose Transfer (DPT) \cite{Neverova2018}. 
The qualitative results are shown in Fig.~\ref{fig:soa}. 
It can be observed that our results show higher realism and better preserve  identity and garment details compared to the other % state-of-the-art
methods. 

\begin{figure}[t]
    \includegraphics[width=1\linewidth]{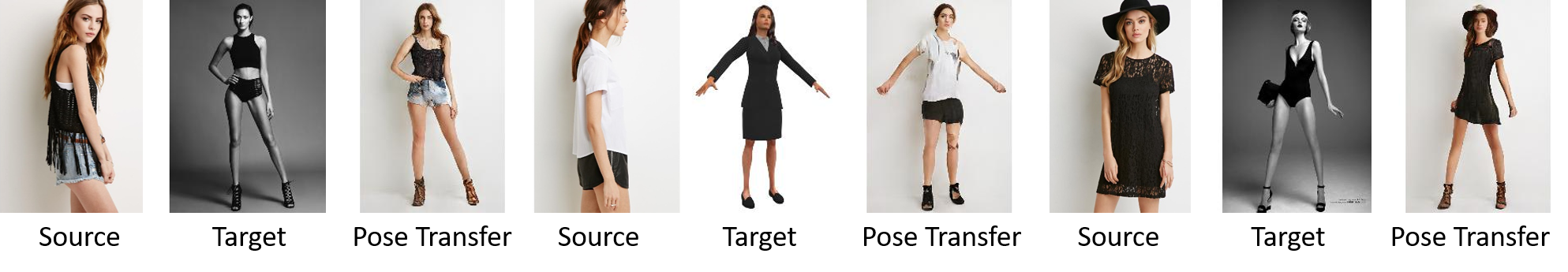}
    \caption{Generalisation of our method to new body poses. The images of the target pose are obtained from the internet.} 
    \label{fig:extreme_pose} 
    \vspace{-0.2cm}
\end{figure}

The quantitative results are provided in Table \ref{table:soa}. Due to inconsistent reporting (or unavailability) of the metrics for the existing approaches, we computed them ourselves. To this end, we collected the results of 176 testing pairs for each state-of-the-art method (the testing pairs and results were kindly provided by the authors of CBI \cite{Grigorev2019CoordinateBasedTI}) and used them for this report. We use the following two metrics for comparison, \textit{i.e.,} 1) Structural Similarity Index (SSIM) \cite{ssim2004} 2) Learned Perceptual Image Patch Similarity (LPIPS)  \cite{zhang2018perceptual}.
SSIM is a structure preservation metric widely used in the existing literature. Though it is an excellent metric for assessment of image degradation quality, it often does not reflect human perception \cite{zhang2018perceptual}. On the other hand, the recently introduced LPIPS claims to capture human judgment better than existing hand-designed metrics. In terms of SSIM, we perform as well as the existing methods, whereas we significantly outperform them on LPIPS metric. 
Please note that similar to other learning-based methods, our approach will struggle with poses that are far from those seen in the training set. However, our method performs well in such scenarios for many cases. Qualitative results on some target poses outside of training dataset distribution are shown in Fig.~\ref{fig:extreme_pose}.

\begin{table}[t]
\caption{Comparison with state-of-the-art methods using various perceptual metrics, Structural Similarity Index (SSIM) \cite{ssim2004} and Learned Perceptual Image Patch Similarity (LPIPS) \cite{zhang2018perceptual}. $\uparrow$ ($\downarrow$) means higher (lower) is better.}
\centering
\begin{tabular}{r|R{2cm}R{2cm}} 
              & SSIM $\uparrow$   & LPIPS  $\downarrow$ \\ \hline
CBI  \cite{Grigorev2019CoordinateBasedTI}          &  0.766     &   0.178    \\
DSC \cite{Siarohin2019AppearanceAP}          &  0.750        &   0.214    \\
VUnet  \cite{esser2018variational}       &  0.739      &    0.202   \\
DPT \cite{Neverova2018}           &   0.759      &     0.206  \\
\textbf{Ours} & \textbf{0.768}  &  \textbf{0.164}   \\ \hline
GT  & 1.0 &  0.0  \\ 
\end{tabular}
\label{table:soa}
\vspace{-0.3cm}
\end{table}

\subsection{User Study} 
To assess the qualitative impact of our method, we perform an extensive user study which compares it with two other state-of-the-art pose transfer methods --  Coordinate Base Inpainting (CBI)~\cite{Grigorev2019CoordinateBasedTI} and DensePose Transfer (DPT)~\cite{Neverova2018}. We train on the DeepFashion dataset \cite{Liu2016DeepFashion} and generate renderings on the test split. 
The user study follows several criteria. 
% is designed following several criteria. 
% 
First, it covers as large a variety of source and target poses.
% as possible. 
Second, the ratio between the male and female samples reflects the same ratio of the dataset. 
It also contains failure cases as those shown in Fig.~\ref{fig:failure_cases} with difficult decisions. 
In total, we prepare $26$ samples containing the source image (explicitly marked as such) and three novel views generated by CBI, DPT and our method (labeled as view A, B or C in randomised order). 
For each sample, two questions are asked: 1) Which view looks the most like the person in the source image? and 2) Which view looks the most realistic? 

The user study was performed with a browser interface, the order of questions is randomised, and $46$ anonymous participants submitted their answers. 
% 
% to avoid the bias. 
% 
The results are as follows. 
The first question has been answered in $46.2\%$ of the cases in favour of CBI, and in $53.8\%$ of the cases in favour of our method. 
In all cases, DPT has always been the last choice. 
The second question has been answered by $30.8\%$  of the participants in favour of CBI, and by $69.2\%$ of the participants in favour of our approach. 
Again, DPT was preferred in no case. 

The user study shows that our method achieves state-of-the-art quality in preserving the identity, and significantly outperforms the baselines in the realism of the generated images. 
In $23\%$ of the overall cases, the participants have preferred CBI as the best identity-preserving method and, at the same time, our method as those producing the most realistic renderings. 
In contrast, there was only one case ($3.8\%$) when our method had been voted as the best identity-preserving and, at the same time, CBI was chosen as the approach producing most realistic renderings. 

\begin{figure}[t] 
    \includegraphics[width=\linewidth]{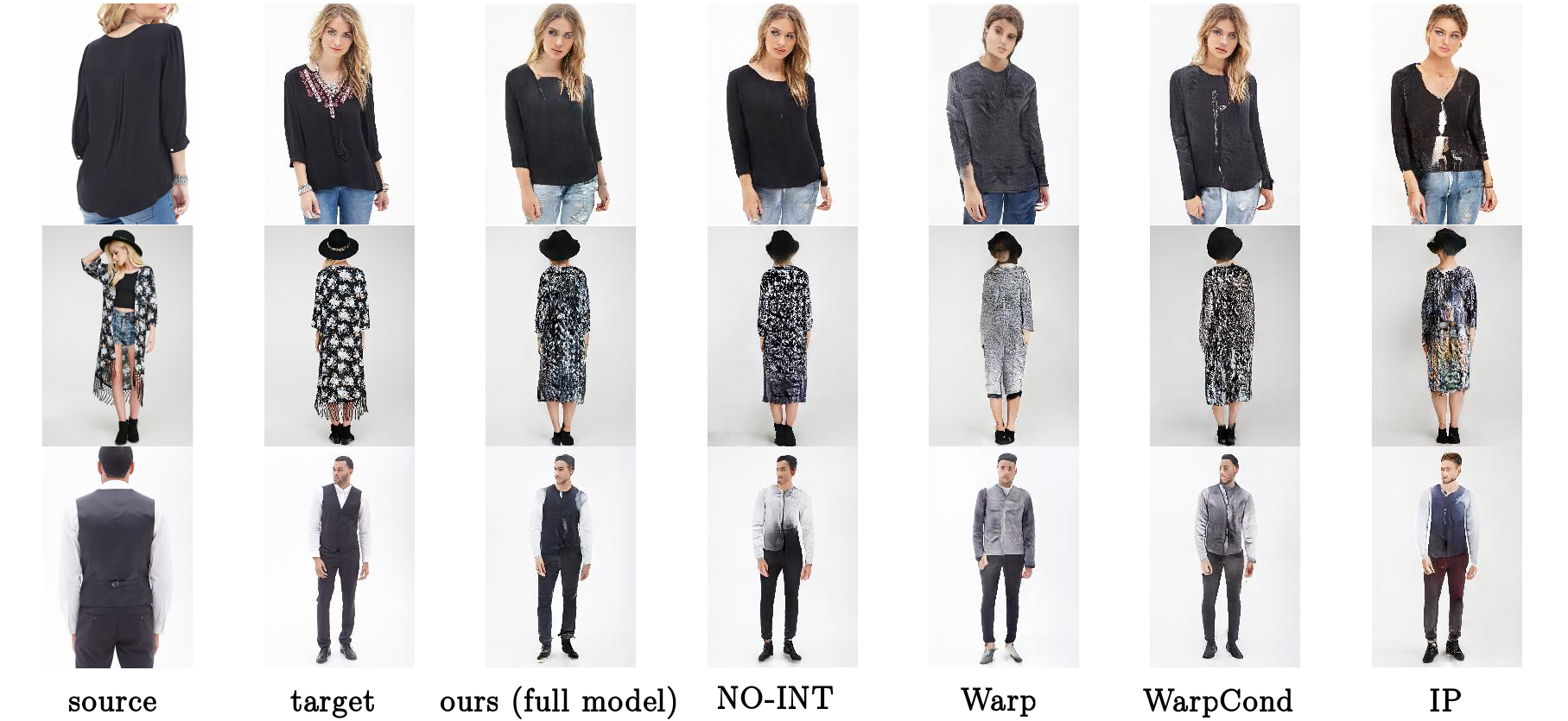} 
    \caption{Results of different baselines and our full model. 
    \textit{No-Int} has no intermediate loss, \textit{Warp} and \textit{WarpCond} perform translation on warped partial texture and \textit{IP} inpaints full colour-texture followed by translation. Under extreme poses and strong occlusions, our method outperforms all the baselines (see Sec.~\ref{sec:ablation}).} 
    \label{fig:ablation} 
    %\vspace{-0.2cm}
\end{figure} 

\subsection{Ablation Study} 
\label{sec:ablation}
To study the advantage of the learned neural texture over other natural choices of texture-based human re-rendering, we created the following three baselines. 

\paragraph{\textbf{IP.}} This baseline involves two stages. First, we train an inpainting network to generate the full UV texture map from the partial UV texture map extracted from the input image. We use the same in-painting loss function as described in Section \ref{sec:lossfunctions} for training this network. After the convergence, we fix and use this network to generate full colour texture from a partial input texture. This full 3-channel UV texture map is then rendered into an intermediate image, and translated through a trained RenderNet $g$. 

\paragraph{\textbf{Warp.}} In this experiment, we warp the incomplete partial UV texture map to the target pose. The reposed incomplete 3-channel intermediate image is then fed to a trained RenderNet $g$ to produce realistic output.

\paragraph{\textbf{WarpCond.}} In this experiment, we warp the partial source texture to the target pose $P_T$ similar to the previous experiment. In addition to the reposed incomplete texture, we also condition the generator network $g$ with the target DensePose image. The target DensePose image acts as a cue to the  generator when the texture information is missing. 

\begin{figure}[t]
    \includegraphics[width=\linewidth]{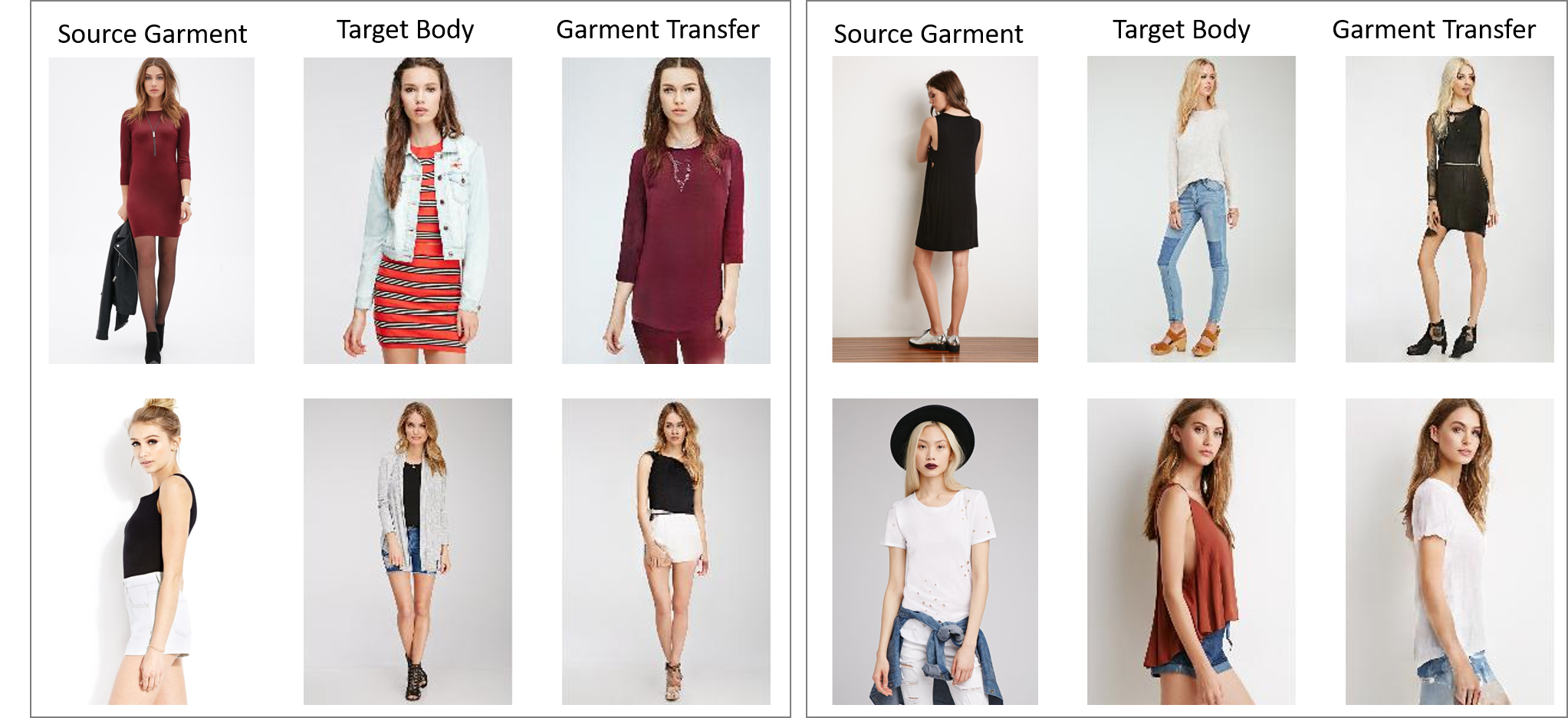}
    \caption{Garment Transfer: Our approach can also be used to render garments from a source image onto the person in a target image.} 
    \label{fig:garment} 
    \vspace{-0.2cm}
\end{figure}

In all these baselines, the architecture of RenderNet $g$ and the losses on the generated image are the same as ours. The only difference is in the number of input channels to RenderNet. Besides, we perform an ablation experiment with an identical pipeline as ours, except we do not enforce any intermediate texture loss, denoted as  \textbf{No-Int}. The qualitative results of all the networks are shown in Fig.~\ref{fig:ablation}. It can be seen that our methods using a richer intermediate representation (\textit{full} and \textit{No-Int}) produce more realistic images than the other baselines. \textit{Baseline-IP} performs well but produces smooth output compared to the other methods. Because of the lack of details, \textit{Baseline-Warp} often produces non-realistic output in both face and garment regions. When the incomplete texture information is supervised with additional DesnsePose image (as in \textit{Baseline-WarpCond}), the output is of higher quality. In the presence of strong occlusions, the method fails, as the translation network $g$ is incapable of performing both inpainting and realistic rendering at the same time. In contrast, our methods performed well in all the scenarios. 
Adding the intermediate texture loss (to mimic real texture) to the part of neural texture helps our network to converge faster. However, over a large number of iterations, the quality of the final result without such intermediate loss (\textit{No-Int}) is similar to that with intermediate loss.

\subsection{Garment Transfer}
Our method can be naturally extended to perform garment transfer without % the need of 
any further training. 
Given an image of a person with the source body, we extract the partial texture $T_s'$ of the body regions (\textit{e.g.,} face, hands and regions with garments which remain unchanged). 
% where we do not want garment appearance to change). 
% 
We use part indices provided by DensePose to extract the partial texture of the required body parts. 
Next, we extract the partial texture $T_t'$ of the `garment regions' of an image with the desired target garments.
We make a union of the extracted partial textures $T_s'\cup T_t'$ based on their texel regions and feed it to our pipeline with the pose $P_s$ of the body image. Note that texel occupancies of $T_s'$ and $T_t'$ are mutually  exclusive as they are extracted from different body parts. 
See Fig.~\ref{fig:garment} for the qualitative results.

\subsection{Motion Transfer} 
\begin{figure}[t]
    \centering
    \includegraphics[width=0.9\linewidth]{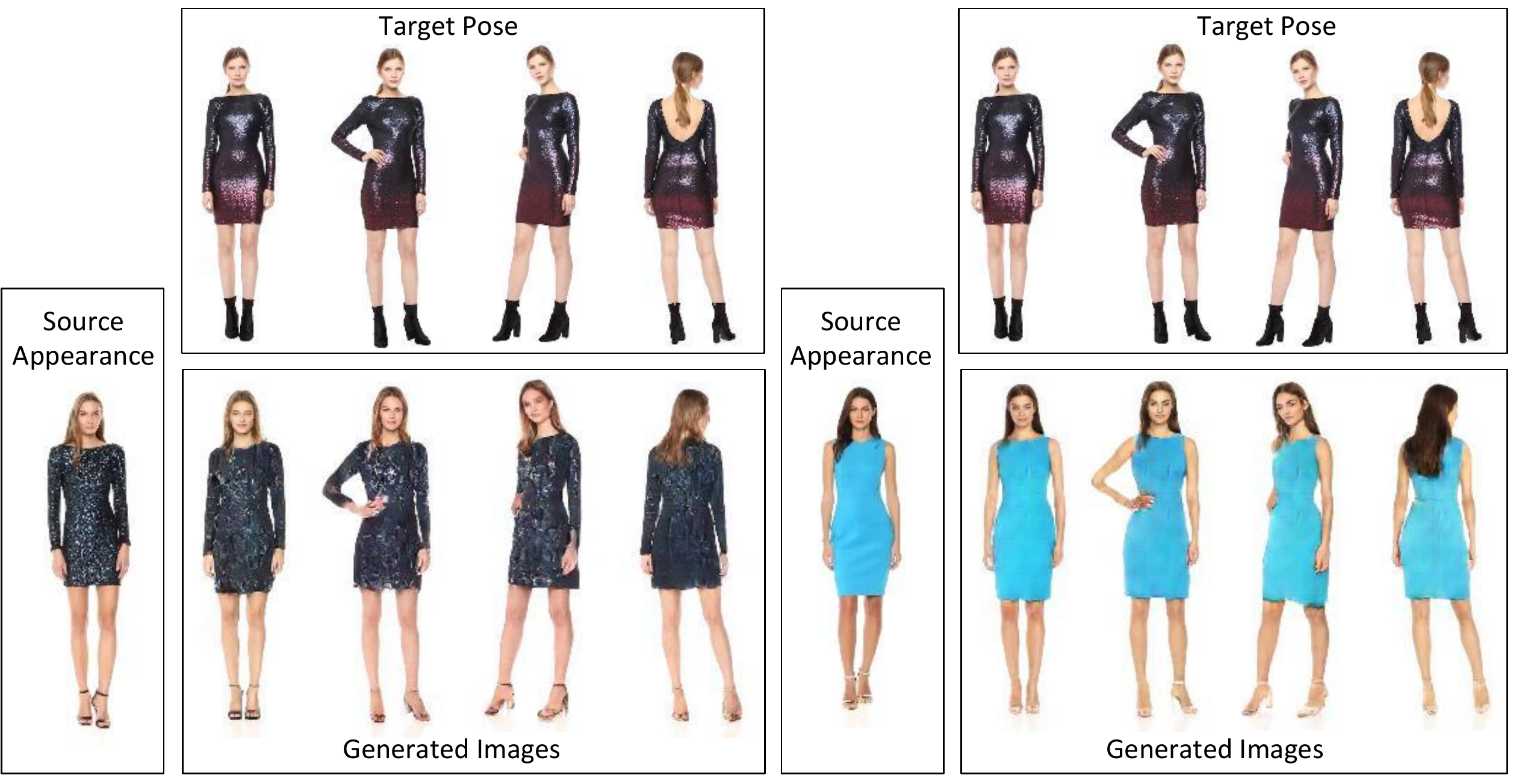}
    \caption{Motion transfer on the Fashion dataset \cite{Zablotskaia2019DwNetDW}. Our approach also can generate realistic renderings for a sequence of poses given a single source image. 
    % Here, results are shown on the .
    } 
    \label{fig:fashion} 
    \vspace{-0.2cm}
\end{figure}

Even though we did not train specifically for generating videos, our method can be applied to each frame of a driving video to create motion transfer. To this end, we keep the source image of the imitator fixed and use the pose from the actor of the driving video (for each frame) in our system to create image animation. We perform the experiment on Fashion Dataset \cite{Zablotskaia2019DwNetDW} and show our results in Fig.~\ref{fig:fashion}.

\begin{figure}[h] 
    \includegraphics[width=\linewidth]{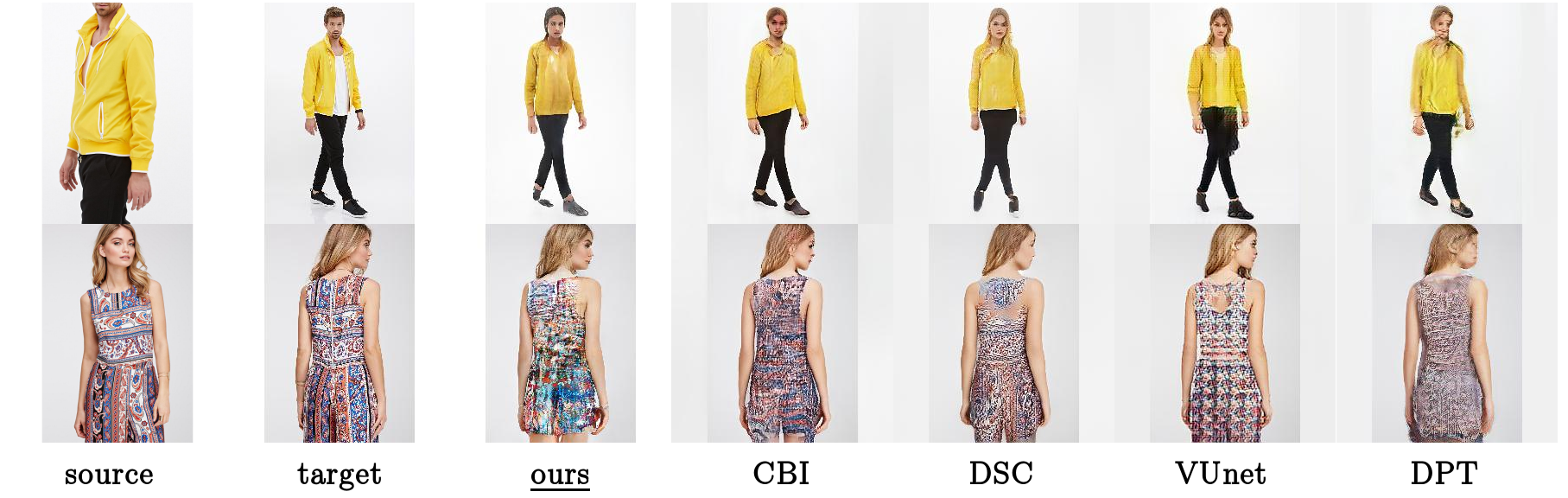} 
    \caption{Limitations: Even though our method produces better quality results than all competing approaches, it nevertheless has some limitations, which are also shared by all competing methods. The top row highlights failures arising out of biases in the training set, while the bottom row highlights failures owing to fine scaled textures which are not effectively captured by any approach. } 
    \label{fig:failure_cases} 
    \vspace{-0.2cm}
\end{figure}

Please refer to the Appendix~\ref{appendix} and the accompanying video for more results. 
% 

%!TEX root = ../article.tex
\section{Discussion}
\label{sec:discuss}

\paragraph{Limitations.} 
Even though we produce high-quality novel views which are preserving the identity and look very realistic, there remain certain limitations for future work to address. 
Fig.~\ref{fig:failure_cases} visualises two representative examples which are difficult for our as well as competing methods. 
In the first row, the head in the source image is only partially visible  
% the source image has the missing head, 
so that the methods have high uncertainty in the frontal facial view and change the gender to female (\textit{e.g.,} hallucinate long hair). 
In the second case, the source texture is too fine-grained for the methods so that some hallucinate repeating patterns and the other ones generate patterns reminiscent of noise. 
In this case, our method generates a texture which is neither repetitive nor looks like noise, and which is still far from the reference. 

\paragraph{Future Extension of Our UV Feature Maps.} 
Instead of sampling RGB textures from the input image to construct a partial UV texture map, learned CNN based features could be used to construct a more informative partial UV feature map, which putatively captures off-geometry details not modelled by the SMPL mesh. Then FeatureNet would convert this partial UV feature map to a full UV feature map. Another alternative would be to use displacement map prediction similar to prior work~\cite{Lazova2019360DegreeTO,alldieck2019tex2shape} to capture off-geometry details.

\section{Conclusion}
\label{sec:conclusion}
In this work, we present an approach for human image synthesis, which allows us to change the camera view and the pose and garments of the subject in the source image.
Our approach uses a high-dimensional UV feature map to encode appearance as an intermediate representation, which is then re-posed and translated using a generator network to achieve realistic rendering of the input subject in a novel pose.
We qualitatively and quantitatively demonstrate the efficacy of our proposed approach at better preserving identity and garment details compared to the other competing methods. Our system, trained once for pose-guided image synthesis, can be directly used for other tasks such as garment and motion transfer. %We conclude that such high-dimensional UV feature-map can be   learned efficiently for the interpretation of a generator network to create realistic images.
%In future work, we plan to jointly learn the geometry from the images in order to provide 
%I think future work can be left out of the conclusion, perhaps ? ok it is already in the discussion 

\section*{Acknowledgement} 

This work was supported by the ERC Consolidator Grant 4DReply (770784). 
% and 

\clearpage
% ---- Bibliography ----
%
% BibTeX users should specify bibliography style 'splncs04'.
% References will then be sorted and formatted in the correct style.
%
\bibliographystyle{splncs04}
\bibliography{article}

\begin{thebibliography}{10}
\providecommand{\url}[1]{\texttt{#1}}
\providecommand{\urlprefix}{URL }
\providecommand{\doi}[1]{https://doi.org/#1}

\bibitem{Agarwal2009C_ACM}
Agarwal, S., Furukawa, Y., Snavely, N., Simon, I., Curless, B., Seitz, S.M.,
  Szeliski, R.: Building rome in a day. Communications of the ACM
  \textbf{54}(10),  105–112 (2011)

\bibitem{alldieck2019tex2shape}
Alldieck, T., Pons-Moll, G., Theobalt, C., Magnor, M.: Tex2shape: Detailed full
  human body geometry from a single image. In: International Conference on
  Computer Vision ({ICCV}) (2019)

\bibitem{Balakrishnan2018}
Balakrishnan, G., Zhao, A., Dalca, A.V., Durand, F., Guttag, J.V.: Synthesizing
  images of humans in unseen poses. Computer Vision and Pattern Recognition
  (CVPR)  (2018)

\bibitem{Buehler2001}
Buehler, C., Bosse, M., McMillan, L., Gortler, S.J., Cohen, M.F.: Unstructured
  lumigraph rendering. In: SIGGRAPH (2001)

\bibitem{Carceroni2002}
Carceroni, R.L., Kutulakos, K.N.: Multi-view scene capture by surfel sampling:
  From video streams to non-rigid 3d motion, shape and reflectance.
  International Journal of Computer Vision (IJCV)  \textbf{49}(2),  175--214
  (2002)

\bibitem{chan2019dance}
Chan, C., Ginosar, S., Zhou, T., Efros, A.A.: Everybody dance now. In:
  International Conference on Computer Vision (ICCV) (2019)

\bibitem{Chaurasia2013}
Chaurasia, G., Duch\^ene, S., Sorkine-Hornung, O., Drettakis, G.: Depth
  synthesis and local warps for plausible image-based navigation. ACM
  Transactions on Graphics  \textbf{32} (2013)

\bibitem{Debevec1998}
Debevec, P., Yu, Y., Borshukov, G.: Efficient view-dependent image-based
  rendering with projective texture-mapping. Eurographics Workshop on Rendering
   (1998)

\bibitem{Dou2016}
Dou, M., Khamis, S., Degtyarev, Y., Davidson, P., Fanello, S.R., Kowdle, A.,
  Escolano, S.O., Rhemann, C., Kim, D., Taylor, J., et~al.: Fusion4d: Real-time
  performance capture of challenging scenes. ACM Trans. Graph.  \textbf{35}(4)
  (2016)

\bibitem{esser2018variational}
Esser, P., Sutter, E., Ommer, B.: A variational u-net for conditional
  appearance and shape generation. In: Computer Vision and Pattern Recognition
  (CVPR). pp. 8857--8866 (2018)

\bibitem{Gortler1996}
Gortler, S.J., Grzeszczuk, R., Szeliski, R., Cohen, M.F.: The lumigraph. In:
  SIGGRAPH. p. 43–54 (1996)

\bibitem{Grigorev2019CoordinateBasedTI}
Grigor'ev, A.K., Sevastopolsky, A., Vakhitov, A., Lempitsky, V.S.:
  Coordinate-based texture inpainting for pose-guided human image generation.
  Computer Vision and Pattern Recognition (CVPR) pp. 12127--12136 (2019)

\bibitem{Guo2017}
Guo, K., Xu, F., Yu, T., Liu, X., Dai, Q., Liu, Y.: Real-time geometry, albedo,
  and motion reconstruction using a single rgb-d camera. ACM Trans. Graph.
  \textbf{36}(4) (2017)

\bibitem{Han_2019_ICCV}
Han, X., Hu, X., Huang, W., Scott, M.R.: Clothflow: A flow-based model for
  clothed person generation. In: Proceedings of the IEEE/CVF International
  Conference on Computer Vision (ICCV) (October 2019)

\bibitem{Huang2018}
Huang, Z., Li, T., Chen, W., Zhao, Y., Xing, J., Legendre, C., Luo, L., Ma, C.,
  Li, H.: Deep volumetric video from very sparse multi-view performance
  capture. In: European Conference on Computer Vision (ECCV). pp. 351--369
  (2018)

\bibitem{johnson2016perceptual}
Johnson, J., Alahi, A., Fei-Fei, L.: Perceptual losses for real-time style
  transfer and super-resolution. In: European Conference on Computer Vision
  (ECCV). pp. 694--711 (2016)

\bibitem{hmrKanazawa17}
Kanazawa, A., Black, M.J., Jacobs, D.W., Malik, J.: End-to-end recovery of
  human shape and pose. In: Computer Vision and Pattern Regognition (CVPR)
  (2018)

\bibitem{Kim2019Neural}
Kim, H., Elgharib, M., Zoll{\"o}fer, Michael~Seidel, H.P., Beeler, T.,
  Richardt, C., Theobalt, C.: Neural style-preserving visual dubbing. ACM
  Transactions on Graphics (TOG)  \textbf{38}(6),  178:1--13 (2019)

\bibitem{Kim2018}
Kim, H., Garrido, P., Tewari, A., Xu, W., Thies, J., Nie{\ss}ner, M.,
  P{\'e}rez, P., Richardt, C., Zoll{\"o}fer, M., Theobalt, C.: Deep video
  portraits. ACM Transactions on Graphics (TOG)  \textbf{37} (2018)

\bibitem{adam}
Kingma, D.P., Ba, J.: Adam: A method for stochastic optimization. In:
  International Conference on Learning Representations (ICLR) (2015)

\bibitem{Lazova2019360DegreeTO}
Lazova, V., Insafutdinov, E., Pons-Moll, G.: 360-degree textures of people in
  clothing from a single image. International Conference on 3D Vision (3DV) pp.
  643--653 (2019)

\bibitem{Levoy1996}
Levoy, M., Hanrahan, P.: Light field rendering. In: SIGGRAPH. p. 31–42 (1996)

\bibitem{Liu2019}
Liu, L., Xu, W., Zollhoefer, M., Kim, H., Bernard, F., Habermann, M., Wang, W.,
  Theobalt, C.: Neural rendering and reenactment of human actor videos. ACM
  Transactions on Graphics (TOG)  (2019)

\bibitem{liu2017sphereface}
Liu, W., Wen, Y., Yu, Z., Li, M., Raj, B., Song, L.: Sphereface: Deep
  hypersphere embedding for face recognition. In: Computer Vision and Pattern
  Recognition (CVPR). pp. 212--220 (2017)

\bibitem{lwb2019}
Liu, W., Piao, Z., Jie, M., Luo, W., Ma, L., Gao, S.: Liquid warping gan: A
  unified framework for human motion imitation, appearance transfer and novel
  view synthesis. In: International Conference on Computer Vision (ICCV) (2019)

\bibitem{Liu2010TVCG}
{Liu}, Y., {Dai}, Q., {Xu}, W.: A point-cloud-based multiview stereo algorithm
  for free-viewpoint video. IEEE Transactions on Visualization and Computer
  Graphics (TVCG)  \textbf{16}(3),  407--418 (2010)

\bibitem{Liu2016DeepFashion}
{Liu}, Z., {Luo}, P., {Qiu}, S., {Wang}, X., {Tang}, X.: Deepfashion: Powering
  robust clothes recognition and retrieval with rich annotations. In: Computer
  Vision and Pattern Recognition (CVPR). pp. 1096--1104 (2016)

\bibitem{lombardi2019neural}
Lombardi, S., Simon, T., Saragih, J., Schwartz, G., Lehrmann, A., Sheikh, Y.:
  Neural volumes: Learning dynamic renderable volumes from images. ACM Trans.
  Graph. (SIGGRAPH)  \textbf{38}(4) (2019)

\bibitem{SMPL:2015}
Loper, M., Mahmood, N., Romero, J., Pons-Moll, G., Black, M.J.: {SMPL}: A
  skinned multi-person linear model. ACM Trans. Graphics (Proc. SIGGRAPH Asia)
  \textbf{34}(6),  248:1--248:16 (Oct 2015)

\bibitem{Ma18}
Ma, L., Sun, Q., Georgoulis, S., van Gool, L., Schiele, B., Fritz, M.:
  Disentangled person image generation. Computer Vision and Pattern Recognition
  (CVPR)  (2018)

\bibitem{MartinBrualla2018}
Martin~Brualla, R., Lincoln, P., Kowdle, A., Rhemann, C., Goldman, D., Keskin,
  C., Seitz, S., Izadi, S., Fanello, S., Pandey, R., Yang, S., Pidlypenskyi,
  P., Taylor, J., Valentin, J., Khamis, S., Davidson, P., Tkach, A.:
  Lookingood: Enhancing performance capture with real-time neural re-rendering.
  ACM Transactions on Graphics (TOG)  \textbf{37} (2018)

\bibitem{Matsuyama2004}
{Matsuyama}, T., {Xiaojun Wu}, {Takai}, T., {Wada}, T.: Real-time dynamic 3-d
  object shape reconstruction and high-fidelity texture mapping for 3-d video.
  IEEE Transactions on Circuits and Systems for Video Technology
  \textbf{14}(3),  357--369 (2004)

\bibitem{Neverova2018}
Neverova, N., G\"{u}ler, R.A., Kokkinos, I.: Dense pose transfer. European
  Conference on Computer Vision (ECCV)  (2018)

\bibitem{OrtsEscolano2016}
Orts-Escolano, S., Rhemann, C., Fanello, S., Chang, W., Kowdle, A., Degtyarev,
  Y., Kim, D., Davidson, P.L., Khamis, S., Dou, M., et~al.: Holoportation:
  Virtual 3d teleportation in real-time. In: Annual Symposium on User Interface
  Software and Technology. p. 741–754 (2016)

\bibitem{Pandey2019}
Pandey, R., Tkach, A., Yang, S., Pidlypenskyi, P., Taylor, J., Martin-Brualla,
  R., Tagliasacchi, A., Papandreou, G., Davidson, P., Keskin, C., Izadi, S.,
  Fanello, S.: Volumetric capture of humans with a single rgbd camera via
  semi-parametric learning. In: Computer Vision and Pattern Recognition (CVPR)
  (2019)

\bibitem{Pfister2000}
Pfister, H., Zwicker, M., van Baar, J., Gross, M.: Surfels: Surface elements as
  rendering primitives. In: SIGGRAPH. p. 335–342 (2000)

\bibitem{Guler2018DensePose}
Rieza Alp~Gueler, Natalia~Neverova, I.K.: Densepose: Dense human pose
  estimation in the wild. In: Computer Vision and Pattern Recognition (CVPR)
  (2018)

\bibitem{RFB15a}
Ronneberger, O., P.Fischer, Brox, T.: U-net: Convolutional networks for
  biomedical image segmentation. In: Medical Image Computing and
  Computer-Assisted Intervention (MICCAI). pp. 234--241 (2015)

\bibitem{saito2019pifu}
Saito, S., Huang, Z., Natsume, R., Morishima, S., Kanazawa, A., Li, H.: Pifu:
  Pixel-aligned implicit function for high-resolution clothed human
  digitization. International Conference on Computer Vision (ICCV)  (2019)

\bibitem{schonberger2016structure}
Schonberger, J.L., Frahm, J.M.: Structure-from-motion revisited. In: Computer
  Vision and Pattern Recognition (CVPR). pp. 4104--4113 (2016)

\bibitem{Shade1998}
Shade, J., Gortler, S., He, L.w., Szeliski, R.: Layered depth images. In:
  SIGGRAPH. p. 231–242 (1998)

\bibitem{Shysheya2019}
Shysheya, A., Zakharov, E., Aliev, K.A., Bashirov, R., Burkov, E., Iskakov, K.,
  Ivakhnenko, A., Malkov, Y., Pasechnik, I., Ulyanov, D., Vakhitov, A.,
  Lempitsky, V.: Textured neural avatars. In: Computer Vision and Pattern
  Recognition (CVPR) (2019)

\bibitem{Siarohin2019AppearanceAP}
Siarohin, A., Lathuili{\`e}re, S., Sangineto, E., Sebe, N.: Appearance and
  pose-conditioned human image generation using deformable gans. Transactions
  on Pattern Analysis and Machine Intelligence (TPAMI)  (2019)

\bibitem{Siarohin_2019_CVPR}
Siarohin, A., Lathuilière, S., Tulyakov, S., Ricci, E., Sebe, N.: Animating
  arbitrary objects via deep motion transfer. In: Computer Vision and Pattern
  Recognition (CVPR) (2019)

\bibitem{Siarohin_2019_NeurIPS}
Siarohin, A., Lathuilière, S., Tulyakov, S., Ricci, E., Sebe, N.: First order
  motion model for image animation. In: Conference on Neural Information
  Processing Systems (NeurIPS) (2019)

\bibitem{simonyan2014very}
Simonyan, K., Zisserman, A.: Very deep convolutional networks for large-scale
  image recognition. arXiv preprint arXiv:1409.1556  (2014)

\bibitem{sitzmann2019deepvoxels}
Sitzmann, V., Thies, J., Heide, F., Nie{\ss}ner, M., Wetzstein, G.,
  Zollh{\"o}fer, M.: Deepvoxels: Learning persistent 3d feature embeddings. In:
  Computer Vision and Pattern Recognition (CVPR) (2019)

\bibitem{Sitzmann2019}
Sitzmann, V., Zollh{\"o}fer, M., Wetzstein, G.: Scene representation networks:
  Continuous 3d-structure-aware neural scene representations. In: Advances in
  Neural Information Processing Systems (NeurIPS) (2019)

\bibitem{DoubleFusion2018}
Tao, Y., Zheng, Z., Guo, K., Zhao, J., Quionhai, D., Li, H., Pons-Moll, G.,
  Liu, Y.: Doublefusion: Real-time capture of human performance with inner body
  shape from a depth sensor. In: Computer Vision and Pattern Recognition (CVPR)
  (2018)

\bibitem{Thies2019}
Thies, J., Zollh\"{o}fer, M., Nießner, M.: Deferred neural rendering: image
  synthesis using neural textures. ACM Transactions on Graphics (TOG)
  \textbf{38} (2019)

\bibitem{thies2020imageguided}
Thies, J., Zollh{\"o}fer, M., Theobalt, C., Stamminger, M., Nie{\ss}ner, M.:
  Image-guided neural object rendering. In: International Conference on
  Learning Representations (ICLR) (2020)

\bibitem{Tung2009}
Tung, T., Nobuhara, S., Matsuyama, T.: Complete multi-view reconstruction of
  dynamic scenes from probabilistic fusion of narrow and wide baseline stereo.
  In: International Conference on Computer Vision (ICCV). pp. 1709--1716 (2009)

\bibitem{varol17_surreal}
Varol, G., Romero, J., Martin, X., Mahmood, N., Black, M.J., Laptev, I.,
  Schmid, C.: Learning from synthetic humans. In: Computer Vision and Pattern
  Regognition (CVPR) (2017)

\bibitem{wang2018pix2pixHD}
Wang, T.C., Liu, M.Y., Zhu, J.Y., Tao, A., Kautz, J., Catanzaro, B.:
  High-resolution image synthesis and semantic manipulation with conditional
  gans. In: Computer Vision and Pattern Recognition (CVPR) (2018)

\bibitem{Waschbuesch2005}
Waschb{\"u}sch, M., W{\"u}rmlin, S., Cotting, D., Sadlo, F., Gross, M.:
  Scalable 3d video of dynamic scenes. The Visual Computer  \textbf{21}(8),
  629--638 (2005)

\bibitem{xu2019deepviewsynthesis}
Xu, Z., Bi, S., Sunkavalli, K., Hadap, S., Su, H., Ramamoorthi, R.: Deep view
  synthesis from sparse photometric images. ACM Trans. Graph.  \textbf{38}(4),
  76:1--76:13 (2019)

\bibitem{Yu2017}
{Yu}, T., {Guo}, K., {Xu}, F., {Dong}, Y., {Su}, Z., {Zhao}, J., {Li}, J.,
  {Dai}, Q., {Liu}, Y.: Bodyfusion: Real-time capture of human motion and
  surface geometry using a single depth camera. In: International Conference on
  Computer Vision (ICCV). pp. 910--919 (2017)

\bibitem{Yu2019}
Yu, T., Zheng, Z., Zhong, Y., Zhao, J., Dai, Q., Pons-Moll, G., Liu, Y.:
  Simulcap: Single-view human performance capture with cloth simulation. In:
  Computer Vision and Pattern Recognition (CVPR) (2019)

\bibitem{Zablotskaia2019DwNetDW}
Zablotskaia, P., Siarohin, A., Sigal, L., Zhao, B.: Dwnet: Dense warp-based
  network for pose-guided human video generation. In: British Machine Vision
  Conference (BMVC) (2019)

\bibitem{Zhang2003}
Zhang, L., Curless, B., Seitz, S.M.: Spacetime stereo: shape recovery for
  dynamic scenes. In: Computer Vision and Pattern Recognition (CVPR) (2003)

\bibitem{zhang2018perceptual}
Zhang, R., Isola, P., Efros, A.A., Shechtman, E., Wang, O.: The unreasonable
  effectiveness of deep features as a perceptual metric. In: Computer Vision
  and Pattern Recognition (CVPR) (2018)

\bibitem{Zhao2017}
Zhao, B., Wu, X., Cheng, Z.Q., Liu, H., Jie, Z., Feng, J.: Multi-view image
  generation from a single-view. In: ACM International Conference on
  Multimedia. p. 383–391 (2018)

\bibitem{Zhou2019DanceDG}
Zhou, Y., Wang, Z., Fang, C., Bui, T., Berg, T.L.: Dance dance generation:
  Motion transfer for internet videos. In: International Conference on Computer
  Vision Workshops (ICCVW) (2019)

\bibitem{ssim2004}
{Zhou Wang}, {Bovik}, A.C., {Sheikh}, H.R., {Simoncelli}, E.P.: Image quality
  assessment: from error visibility to structural similarity. IEEE Transactions
  on Image Processing  \textbf{13}(4),  600--612 (2004)

\bibitem{Zhu2018extrapolation}
Zhu, H., Su, H., Wang, P., Cao, X., Yang, R.: View extrapolation of human body
  from a single image. In: Computer Vision and Pattern Recognition (CVPR)
  (2018)

\bibitem{Zhu2018}
Zhu, J.Y., Zhang, Z., Zhang, C., Wu, J., Torralba, A., Tenenbaum, J., Freeman,
  B.: Visual object networks: image generation with disentangled 3d
  representations. In: Conference on Neural Information Processing Systems
  (NeurIPS). pp. 118--129 (2018)

\end{thebibliography}

\clearpage
\appendix
\section{Appendix}\label{appendix}
This appendix provides details of the network architectures employed, as well as snippets from the user study. 
% 
% Please refer to the main document and the project  webpage\footnote{\url{http://gvv.mpi-inf.mpg.de/projects/NHRR/}} for further details. 
%We also provide an additional video as a part of the supplementary material of the paper.
%In addition to this document, we also provide a video as a part of the supplementary material. 

%The project details can be found in

\subsection{Network Architecture} 
\subsubsection{FeatureNet.} 
\begin{figure}[t]
    \includegraphics[width=\linewidth]{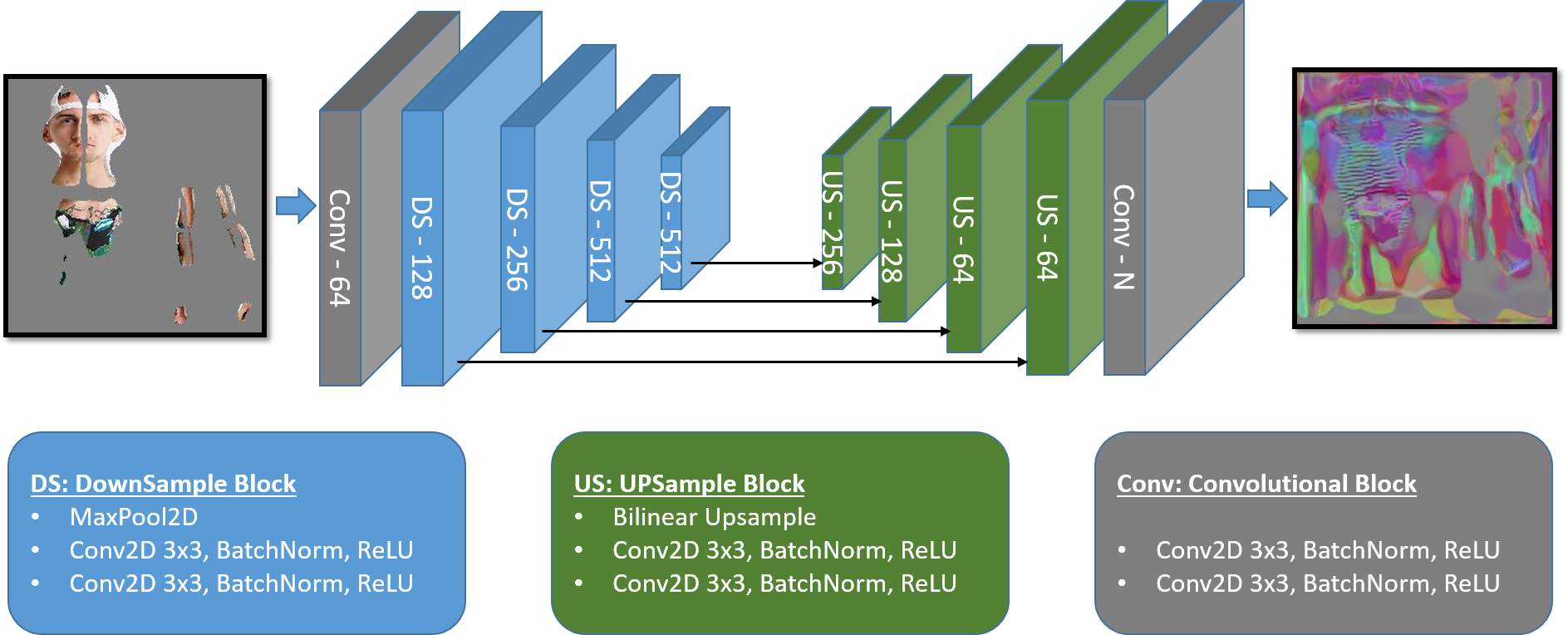}
    \caption{\textbf{FeatureNet ($f$)}: Network 1 of our full pipeline \textit{(Fig. \ref{fig:pipeline})}. FeatureNet converts the partial UV texture map to a full UV feature map, which encodes a richer $N$-dimensional representation at each texel. DS-\textless M\textgreater Denotes a DownSampling block containing MaxPool2D and double convolution with $M$ ouput features. Similar is the case for \textit{US} and \textit{Conv} block.}
    \label{fig:unet} 
\end{figure}

FeatureNet is a U-Net-based network that construct the full \textit{UV feature map} from the partial RGB UV texture map. The network architecture is shown in Fig.~\ref{fig:unet}. 
FeatureNet comprises of four down-sampling blocks followed by four up-sampling blocks. We use the texture of resolution 256 $\times$ 256. 
At the middle most layer the intput is transformed to an activation volume of spatial dimension of 16 $\times$ 16. In all the figures, the 16 dimensional feature image is visualized by projecting it to three dimensions by a fixed random matrix.

\subsubsection{RenderNet.} 
\begin{figure}[h]
    \includegraphics[width=\linewidth]{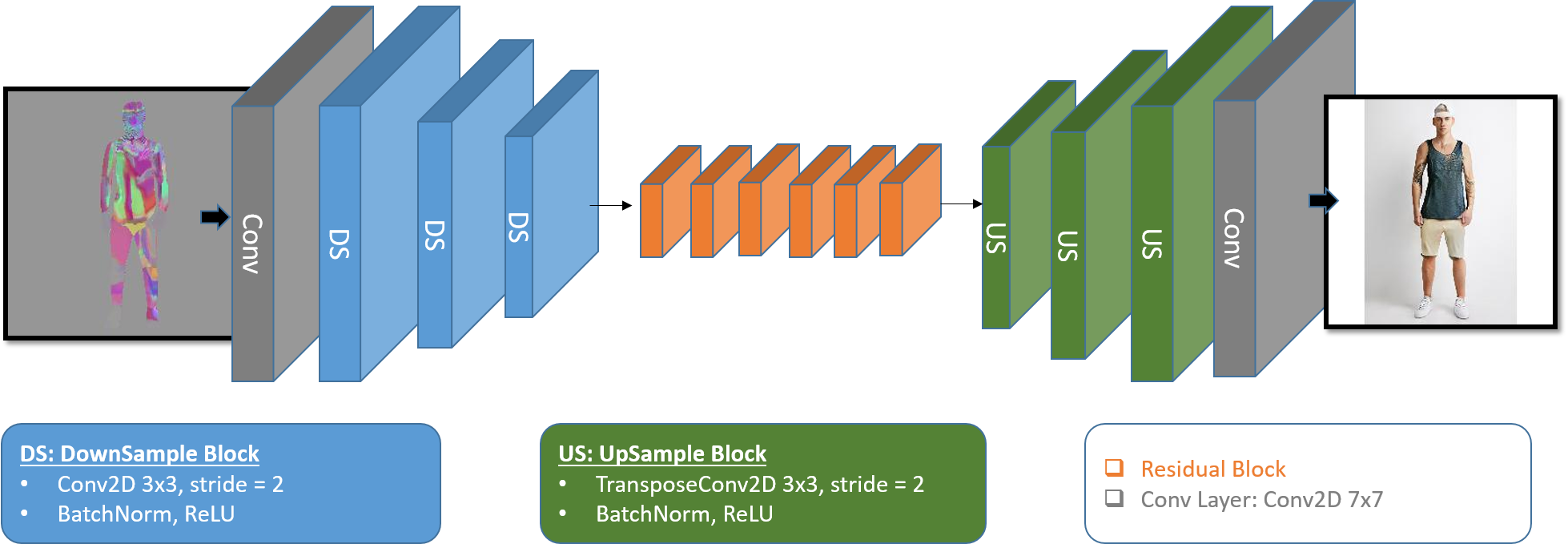}
    \caption{\textbf{RenderNet ($g$)}: Network 2 of our full pipeline \textit{(Fig. \ref{fig:pipeline})}. RenderNet is a translation network that translates the rendered $d$-dimensional \textit{feature image} to a photorealistic image. The network comprises of (a) 3 down-sampling blocks, (b) 6 residual blocks, (c) 3 up-sampling  blocks and finally (d) a convolution layer with \textit{tanh}  activation that gives the final output.} 
    \label{fig:RenderNet} 
\end{figure}

Rendernet translates the rendered \textit{feature image} (from source Feature Map and target pose) to a photorealistic image.
The network architecture is shown in Fig.~\ref{fig:RenderNet}. Both RendereNet and FeatureNet are trained \textit{together} end-to-end following the full pipelilne (Fig. \ref{fig:pipeline}). 
% 

% \section{Network intermediates}
% Intermediate feature activations of an example are shown in Figure \ref{fig:intermediate}.

%\section{Network intermediates}
%Different versions of our networks (Section 4.4 Main paper) different input/output for FeatureNet and RenderNet. Or the \textbf{Baseline-IP}, we use FeatureNet to output 3 dimensional texture image to inpaint the partial texture. In this case, the entire pipeline is executed in stages. Here FeatureNet (with N = 3) is trained untill convergence, which is then fixed for the subsequent training of RenderNet. Inpainting results of this baseline architecture are shown in Figure \ref{fig:inpainting}.

% \begin{figure}[t]
%     \includegraphics[width=\linewidth]{Figures/inpainting.png}
%     \caption{Inpainting results of our \textbf{Baseline-IP}. From left to right - partial texture, inpainted texture, rendering of inpainted texture through target pose, ground-truth target image. We observe that our inpainting network (FeatureNet with 3 channel output) converge quickly by training with image pairs, and provide good quality inpainting results - unlike what is claimed in the existing work \cite{esser2018variational}. After the convergence, we keep the inpainting network fixed and train RenderNet for this baseline.}
%     % 
%     \label{fig:inpainting} 
% \end{figure}

\subsection{Further Results} 
\begin{figure}[t] 
    \includegraphics[width=\linewidth]{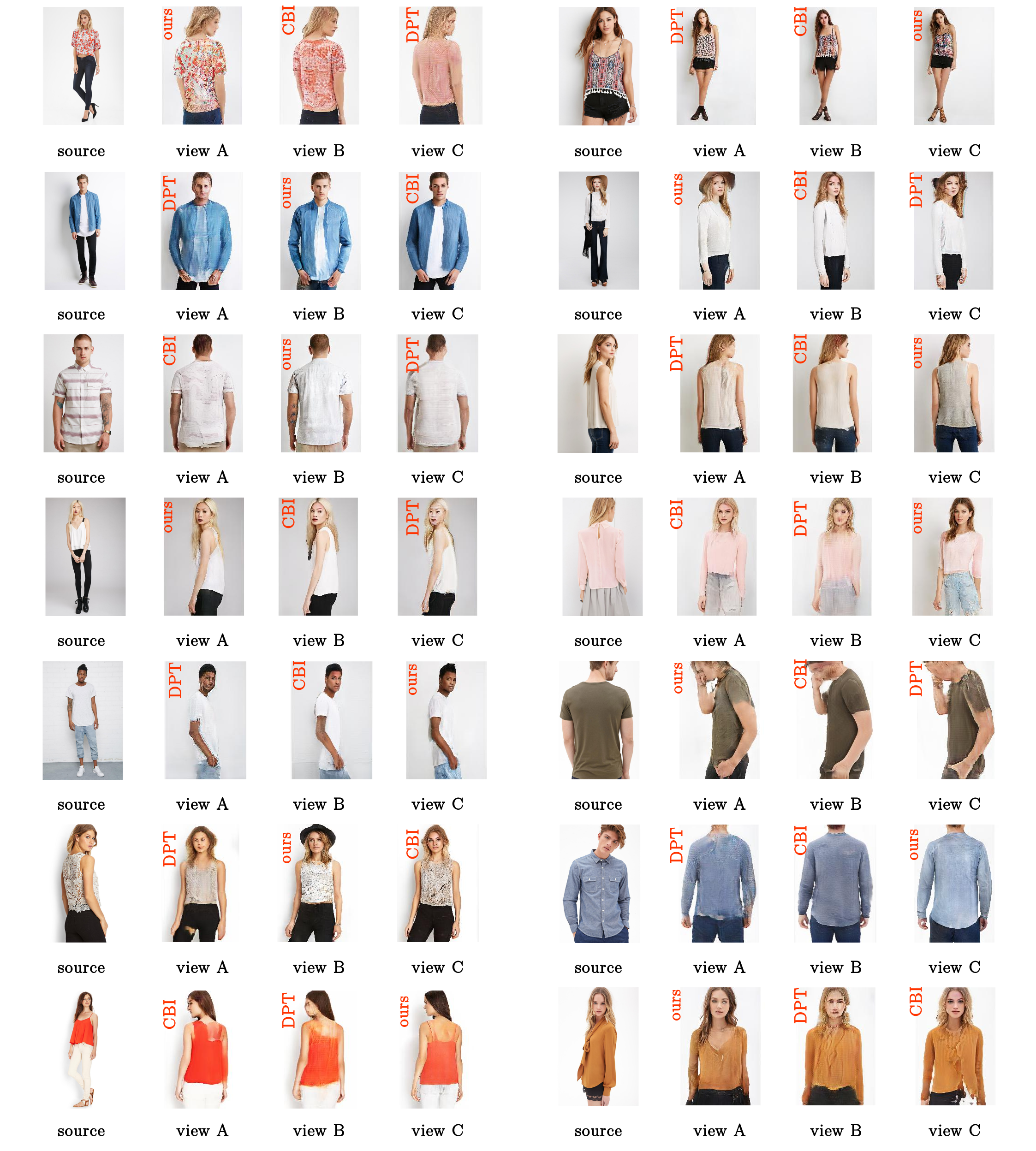} 
    \caption{ 
    The first $14$ samples from the used study (out of $26$). 
    We show the source image and three views generated by CBI  \cite{Grigorev2019CoordinateBasedTI}, DPT \cite{Neverova2018} and our method, in a  randomised order.
    % 
    %(CBI) \cite{Grigorev2019CoordinateBasedTI}, Deformable GAN (DSC) \cite{Siarohin2019AppearanceAP}, Variational U-Net (VUnet) \cite{esser2018variational} and Dense Pose Transfer (DPT) \cite{Neverova2018}.
    % 
    The keys---which were not exposed during the user study---are shown in orange. 
    See Fig.~\ref{fig:user_study_part_2} for the remaining samples. 
    } 
    \label{fig:user_study_part_1} 
\end{figure}

\begin{figure}[t] 
    \includegraphics[width=\linewidth]{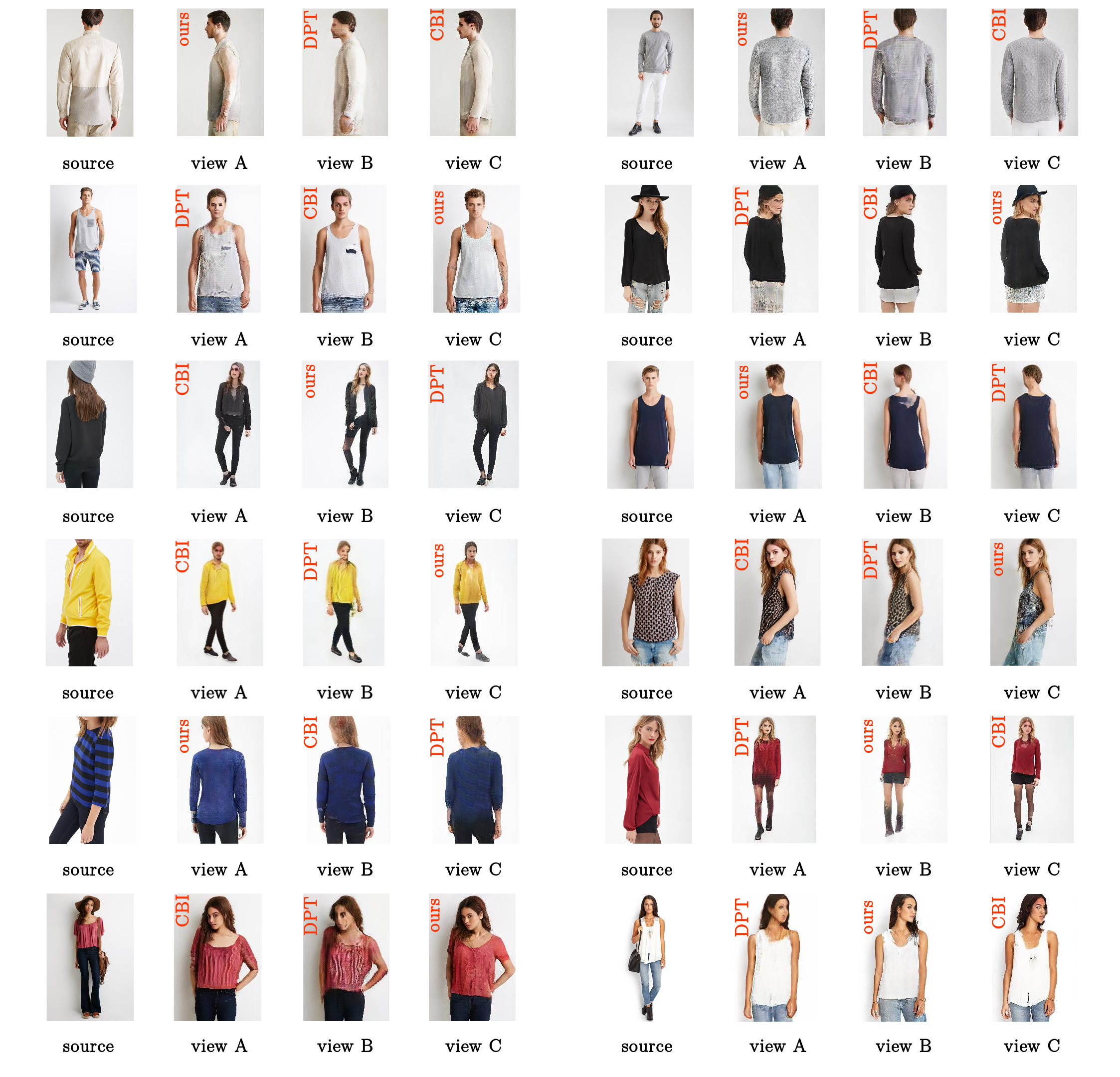} 
    \caption{The further $12$ samples from the used study (out of $26$). } 
    \label{fig:user_study_part_2} 
\end{figure}

In Figs.~\ref{fig:user_study_part_1} and \ref{fig:user_study_part_2} we show our results and its comparison to Coordinate Based Inpainting (CBI)  \cite{Grigorev2019CoordinateBasedTI} and DensePose Transfer (DPT) \cite{Neverova2018}. 
Here, we present the list of the figures that was used in the user study. 

\end{document}